\documentclass[]{spie}

\usepackage{graphicx}
\usepackage{amsmath,amsfonts,amssymb}

\usepackage{subcaption}
\usepackage{hyperref}
\usepackage{booktabs}
\usepackage{float}

\newcommand{\supit}[1]{\textsuperscript{#1}}
\newcommand{\skiplinehalf}{\vspace{0.5\baselineskip}}

\title{Segmentation of Gray Matters and White Matters from Brain MRI data}

\author{Chang Sun\supit{1}, Rui Shi\supit{1}, Tsukasa Koike\supit{2}, Tetsuro Sekine\supit{3}, Akio Morita\supit{4}, and Tetsuya Sakai\supit{1}
\skiplinehalf
\supit{1}Department of Computer Science and Engineering, Waseda University, Shinjuku, Tokyo, Japan \\
\supit{2}Department of Neurosurgery, Teraoka Memorial Hospital, Fukuyama, Hiroshima, Japan \\
\supit{3}Department of Radiology, Nippon Medical School Hospital, Tokyo, Japan \\
\supit{4}Tokyo Rosai Hospital, Ota, Tokyo, Japan
}

\begin{document}

\maketitle

\begin{abstract}
Accurate segmentation of brain tissues such as gray matter and white matter from magnetic resonance imaging is essential for studying brain anatomy, diagnosing neurological disorders, and monitoring disease progression. Traditional methods, such as FSL FAST, produce tissue probability maps but often require task-speciﬁc adjustments and face challenges with diverse imaging conditions. Recent foundation models, such as MedSAM, offer a prompt-based approach that leverages large-scale pretraining. In this paper,\footnote{This is the full-length version of a paper accepted for ICMIP 2026.}\footnote{Note: The ICMIP 2026 proceeding version (Paper ID PP525) contains typographical errors in Table 2. The correct numerical results are presented in this arXiv version. Readers are advised to refer to this version for accurate results.} we propose a modiﬁed MedSAM model designed for multi-class brain tissue segmentation. Our preprocessing pipeline includes skull stripping with FSL BET, tissue probability mapping with FSL FAST, and converting these into 2D axial, sagittal, coronal slices with multi-class labels (background, gray matter, and white matter). We extend MedSAM's mask decoder to three classes, freezing the pre-trained image encoder and ﬁne-tuning the prompt encoder and decoder. Experiments on the IXI dataset achieve Dice scores up to 0.8751. This work demonstrates that foundation models like MedSAM can be adapted for multi-class medical image segmentation with minimal architectural modifications. Our findings suggest that such models can be extended to more diverse medical imaging scenarios in future work.
\end{abstract}


\section{INTRODUCTION}
\label{sec:introduction}

Magnetic Resonance Imaging (MRI) is a non-invasive, high-resolution method of visualizing the internal structure and certain aspects of function within the body. A major advantage is that the MRI scanner is able to differentiate between tissues with subtle signal intensity differences \cite{katti2011magnetic}. Gray matter and white matter play important roles in brain function \cite{barha2016basics}. Gray matter consists of neuronal cell bodies, dendrites, and synapses. White matter is composed of myelinated axons. White matter develops more slowly than gray matter, but also shows more rapid loss in older age \cite{mercadante2020neuroanatomy,mao2018modeling}.

The change in gray matter and white matter also affects human health. For example, heart failure is associated with localized gray matter deterioration. Loss of gray matter can also cause autonomic disturbances, memory deﬁcits, and sleep-disordered breathing \cite{woo2003regional}. Aging can cause a reduced volume, compromised structural integrity of myelinated axons and an increase in white matter hyperintensities in white matter. These changes are closely linked to cognitive decline and neurological disabilities \cite{groh2025white}.

Therefore, white matter and gray matter segmentation are crucial for monitoring human health welfare. There are different methods designed for segmentation: Traditional methods include thresholding, region growing, and K-means.

Deep learning models have also been developed over the years for the segmentation of brain MRI images, such as the U-Net architecture \cite{ronneberger2015u}. Building on the U-Net architecture, nnU-Net framework was proposed \cite{isensee2021nnu}. More recently, foundation models such as MedSAM offer ﬂexible, prompt-based segmentation \cite{ma2024segment}.

Although MedSAM works well for binary segmentation (foreground vs background), its application to multi-class tasks like gray matter and white matter segmentation remains underexplored.

In this paper, we propose a modiﬁed MedSAM model tailored for multi-class tissue segmentation of brain MRI. We extend the original mask decoder to output three classes (background, gray matter, and white matter), while freezing the encoder and ﬁne-tuning the prompt decoder. We evaluate whether this minimally modiﬁed architecture can achieve accurate segmentation on T1-weighted MRI slices from different anatomical orientations.

\section{BACKGROUND AND RELATED WORK}
\label{sec:background and related work}

In this section, we provide the background relevant to brain MRI segmentation, including skull stripping with FSL BET, tissue segmentation with FSL FAST, and fine-tuning approaches in medical imaging.

\subsection{Brain Extraction with FSL BET}

The Brain Extraction Tool (BET), part of the FMRIB (Functional MRI of the Brain) Software Library, is widely used for removing
non-brain tissue (e.g., skull, scalp) from T1-weighted MRI volumes. BET uses a deformable
surface model to estimate the brain boundary. It has become a standard preprocessing step in many
neuroimaging workflows. This helps improve the accuracy of downstream tasks such as
segmentation and registration \cite{smith2002fast}.

\subsection{Tissue Segmentation with FSL Fast}

FSL FAST segments brain tissue into gray matter, white matter, and cerebrospinal fluid by modeling voxel intensities with a hidden Markov random field and applying the Expectation-Maximization (EM) algorithm. It produces probabilistic maps for each of the gray matter, white matter and cerebrospinal fluid. The probabilistic maps are used in many research pipelines as pseudo ground truth \cite{zhang2002segmentation}.

\subsection{Fine-Tuning in Medical Imaging}

Fine-tuning adapts a pre-trained model to a target task by continuing training on domain-specific data. In medical imaging, this is critical due to the limited availability of labeled data. Tajbakhsh et al. (2016) demonstrated that fine-tuned models consistently outperform training-from-scratch models when data is limited, especially in classification and segmentation tasks \cite{tajbakhsh2016convolutional}.

\subsection{Traditional Methods}

There are traditional MRI segmentation methods, such as thresholding, region growing, and K-means. Thresholding is one of the oldest techniques \cite{ahmed2008}. It assumes that images are composed of regions with different gray level ranges. A thresholding procedure determines an intensity value (threshold) that separates the desired classes. Grouping all pixels with intensities between two thresholds into one class completes the segmentation process. Region growing is a technique for extracting a region of an image that is connected based on some predefined criteria \cite{pan2007regiongrowing}. This criterion is determined based on the edges and/or intensities in the image. This requires an operator to select a seed point manually and extract all pixels connected to the initial seed with the same intensity value. As its name suggests, region growing iteratively expands the segmented area. In the next step, we examine the pixels in a small neighborhood region and add them to the growing regions. K-means clustering algorithm is an unsupervised method that classifies input data points into multiple classes based on their inherent distance from each other \cite{abras2005kmeans,ng2006clustering}. The k-means algorithm assumes that the data features form a vector space. Then it tries to find natural clustering in them.

Additional traditional approaches include boundary-based models, such as the parametric/nonparametric
deformable model, which are boundary-based methods, and level set methods, which
are hybrid methods \cite{salem2013review}.

\subsection{Deep Learning Methods}

Apart from the traditional methods that have been mentioned above, this section shares related works of segmentation using different deep learning model structures and training methods \cite{wu2025survey}.

One of the early model architectures introduced is the U-Net architecture by Ronneberger et al. (2015) \cite{ronneberger2015u}. This is a CNN architecture that employs a symmetric encoder-decoder structure with skip connections. The contracting path in this structure is used for capturing context, and the symmetric expanding path is used for precise localization.

Different model architectures have also been introduced. For example, Kamnitsas et al. (2017) introduced DeepMedic, a dual pathway, 11-layer deep 3D CNN, which is able to perform the task of brain lesion segmentation \cite{kamnitsas2017efficient}.

Building on the U-Net architecture, nnU-Net, unlike traditional architectures that require manual design and tuning, is a self-configuring framework that eliminates the need for manual architecture design and training configuration by automatically adapting model parameters based on the characteristics of the dataset \cite{isensee2021nnu}.

\subsection{MedSAM}

MedSAM \cite{ma2024segment} is a foundation model built on the Segment Anything Model (SAM) \cite{kirillov2023segment}. MedSAM adapted the vision transformer architecture and prompt-based segmentation of the original SAM model for medical images such as CT and MRI. Although the original SAM model was developed for general object segmentation, MedSAM extends its capabilities to domain-speciﬁc tasks \cite{ma2024segment}. This study further investigates MedSAM's extension to multi-class medical image segmentation.

Currently, most existing models, such as U-Net and nnU-Net, are task-speciﬁc, where they need to be trained from scratch or re-tuned for each segmentation task. While previous approaches rely on task-speciﬁc training from scratch, our work explores the potential of foundation models like MedSAM, which aim to generalize across segmentation tasks with minimal ﬁne-tuning.

We extend the binary segmentation capacity of MedSAM to multi-class prediction, enabling ﬁner-grained anatomical segmentation without redesigning the entire model. We will provide the details in Section \ref{sec:method}).

\section{METHOD}
\label{sec:method}

In this section, we introduce the details of the proposed method.
We propose a method that is able to label the gray matter, white matter, and background of brain MRI images through a fine-tuned deep-learning foundation model.

\subsection{Modifying MedSAM Model Architecture}

The original MedSAM model was designed for binary segmentation. We modified the core model architecture to achieve multi-class segmentation for brain MRI images.

We modified the mask decoder in the MedSAM architecture to enable segmentation of 3 classes (white matter, gray matter, and background). The original decoder produced a single-channel (binary) mask. We added a new convolutional layer to output 3 channels, one for each class. This allowed the model to generate a separate logit map for each class at every pixel. 

We obtained the final segmentation by taking the argmax across the channel dimension of the output tensor. This assigned each pixel to the class with the highest predicted score.

\begin{figure}[H]
\centering
\includegraphics[width=\textwidth]{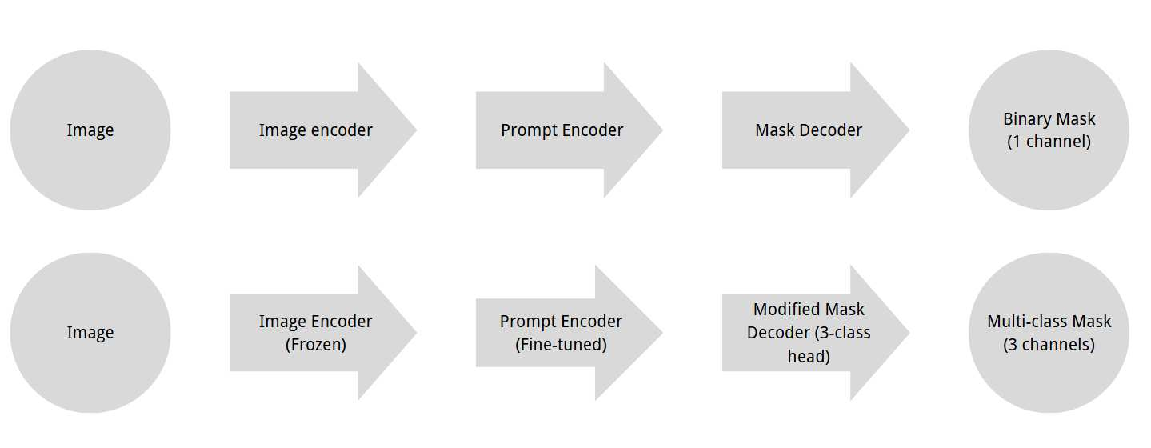}
\caption{Comparison between the original MedSAM model architecture and the modiﬁed model architecture.}
\label{fig:flowchart}
\end{figure}

Figure \ref{fig:flowchart} summarizes how we adapt MedSAM for three-class brain tissue segmentation. The only architectural change is replacing the original mask output head with a three-channel prediction head (BG/GM/WM), while keeping the remainder of the pipeline unchanged. This enables multi-class segmentation while preserving the pretrained representation.

\subsection{Preprocessing Image}

The original MRI datasets consist of 3D brain volumes in NIfTI format. Although the data are available as 3D volumes, we adopt a 2D slice-based approach. This design choice is consistent with the MedSAM framework, which operates on 2D images, and also aligns with the original data handling pipeline based on 2D DICOM slices. Using 2D slices allows efficient training while preserving compatibility with the pretrained model. We developed a multi-stage pipeline to convert these raw volumes into paired 2D image slices and multi-class masks. This pipeline is designed for the segmentation of white matter, gray matter, and background.

\subsubsection{FSL BET and FSL FAST}

First, we applied the FSL BET tool (Brain Extraction Tool) \cite{smith2002fast} to all images in the training, validation, and test sets. This step removes the non-tissue (such as the skull) from the brain MRI volumes. Next, we performed tissue segmentation using FSL FAST \cite{zhang2002segmentation} on the brain-extracted images in the training, validation, and test sets. FSL FAST generates probability maps for different tissue types, including gray matter and white matter.

\subsubsection{Slicing and Multi-Class Mask Generation}

Each preprocessed 3D brain volume was sliced along three orthogonal planes: the \textbf{axial plane} (top-down view), the \textbf{coronal plane} (front-back view), and the \textbf{sagittal plane} (side view). Slicing was performed independently for the training, validation, and test datasets in three orientations.

For each slice, the corresponding gray matter and white matter probability maps (generated by FSL FAST) were extracted and resized to a fixed resolution of \textbf{256 × 256 pixels}. Then the resized probability maps were thresholded and combined into a single multi-class mask, assigning each pixel to one of three classes: background (0), gray matter (1), or white matter (2). We have applied a probability threshold of 0.5, where voxels with a gray matter probability greater than 0.5 were assigned the label 1, and voxels with a white matter probability greater than 0.5 were assigned the label 2. All remaining voxels, such as cerebrospinal fluid (CSF), were labeled 0, representing background. This study focuses specifically on differentiating gray matter and white matter tissues.

Slices with minimal brain tissue content were excluded to reduce noise and ensure meaningful supervision. The resulting image-mask pairs were saved as grayscale and label PNG files. This process produced three complete 2D datasets: one axial, one coronal, and one sagittal.

\subsection{Fine-tuning model}

After modifying the MedSAM architecture to support multi-class segmentation, we fine-tuned the model on three separate datasets constructed from the three orientations. This enabled us to investigate how anatomical orientation affects segmentation performance. We have also fine-tuned the model with the unified dataset (all three datasets combined).

For each orientation, the corresponding dataset was divided into training, validation, and test subsets. We used the same architecture and training settings for both training pipelines to ensure a fair comparison.

We initialized the model with pre-trained weights from MedSAM (ViT-B). We froze the image encoder during training (we only fine-tuned the prompt encoder and mask decoder). This was because we could preserve the general features learned from pre-training. This would have reduced the risk of overfitting the relatively small MRI dataset \cite{raghu2019transfusion,dosovitskiy2020image}. In addition, it required lower memory and computational resources. We only updated the prompt encoder and the modified multi-class mask decoder.

Training was conducted for 10 epochs using the AdamW optimizer \cite{loshchilov2019decoupledweightdecayregularization} with a learning rate of \texttt{1e-4} and a batch size of \texttt{2}. We chose AdamW optimizer for stability and better generalization with weight decay. We used a weighted cross-entropy loss to handle class imbalance. Assigning a lower weight to the background class lets the model focus more on the gray matter and white matter regions. Validation was performed after every epoch, and early stopping was applied based on validation loss.

The best-performing models from axial, coronal, sagittal, and unified training were saved and then evaluated on a common test set to assess the effect of orientation on segmentation quality.

\begin{figure}[H]
\centering
\includegraphics[width=0.5\textwidth]{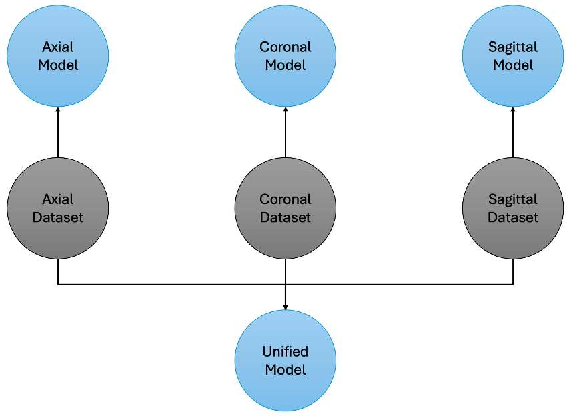}
\caption{Training pipeline comparison: single-orientation models are trained on slices from one anatomical plane (axial, sagittal, or coronal), while the unified model is trained on combined data from all three orientations.}
\label{fig:model_diff}
\end{figure}

Figure \ref{fig:model_diff} illustrates the training setup used in this study. Separate models were trained using slices from individual anatomical orientations (axial, sagittal, coronal), while we trained a unified model using slices from all three orientations combined. This setup allows us to evaluate how anatomical orientation influences segmentation performance.

\section{EXPERIMENTS}

\subsection{Dataset}

We obtained the dataset from the publicly available IXI Dataset (https://brain-development.org/ixidataset/) that contains structural brain MRIs from healthy subjects. Specifically, we used the T1-weighted MRI scans provided in NIfTI format.

The dataset was collected from 3 different hospitals in London:

\begin{itemize}
    \item \textbf{Hammersmith Hospital} – Philips 3T system
    \item \textbf{Guy's Hospital} – Philips 1.5T system
    \item \textbf{Institute of Psychiatry} – GE 1.5T system
\end{itemize}

In total, the dataset includes 581 brain scans. Then we randomly split the dataset into three subsets:

\begin{itemize}
    \item Training set: 406 (70\%)
    \item Validation set: 87 (15\%)
    \item Test set: 88 (15\%)
\end{itemize}

Then we applied the method for preprocessing the data mentioned in the method section to process the data. We prepared two sets of datasets, 1 based on axial slicing and 1 on sagittal slicing. After slicing, we obtained the following number of slices for each dataset and each model.

Axial training slices: 
\begin{itemize}
    \item Training set: 72963 slices
    \item Validation set: 15469 slices
    \item Test set: 15815 slices
\end{itemize}

Sagittal training slices: 
\begin{itemize}
    \item Training set: 45842 slices
    \item Validation set: 9817 slices
    \item Test set: 9885 slices
\end{itemize}

\subsection{Evaluation Metrics}

The goal of our evaluation is to assess how accurately the trained models can segment brain tissue from 2D MRI slices. We evaluate all four models with the same pair of evaluation matrices.

\subsubsection{Dice Similarity Coefficient (DSC)}

Dice measures the overlap between the predicted and ground truth regions. It emphasizes how well the two regions match, especially in cases with class imbalance. It ranges from 0 (no overlap) to 1 (complete overlap). The Dice coefficient between set A and set B can be defined as:

\begin{equation}\label{eq1}
    Dice(A,B) = \frac{2|A \cap B|}{|A|+|B|}
\end{equation}

Where \(A\) is the set of predicted pixels for a class, \(B\) is the set of ground-truth pixels for the same class. \(|A \cap B|\) is the number of pixels correctly predicted as belonging to that class.

\subsubsection{Intersection over Union (IoU)}

IoU evaluates the size of the intersection divided by the size of the union between the predicted and ground-truth masks, providing a direct measure of segmentation accuracy. It is particularly useful for assessing the quality of predictions in object detection and segmentation tasks, as it penalizes both over- and under-segmentation. IoU between A and B can be defined as:

\begin{equation}\label{eq2}
    IoU(A,B) = \frac{|A \cap B|}{|A \cup B|}
\end{equation}

Where the same as above, \(A\) is the set of predicted pixels for a class, \(B\) is the set of ground-truth pixels for the same class. \(|A \cup B|\) is the total number of pixels in either the predicted or ground-truth region.

\section{RESULTS AND DISCUSSIONS}
\label{sec:results and discussions}

\subsection{Results}

\begin{figure}[H]
\centering
\includegraphics[width=\textwidth]{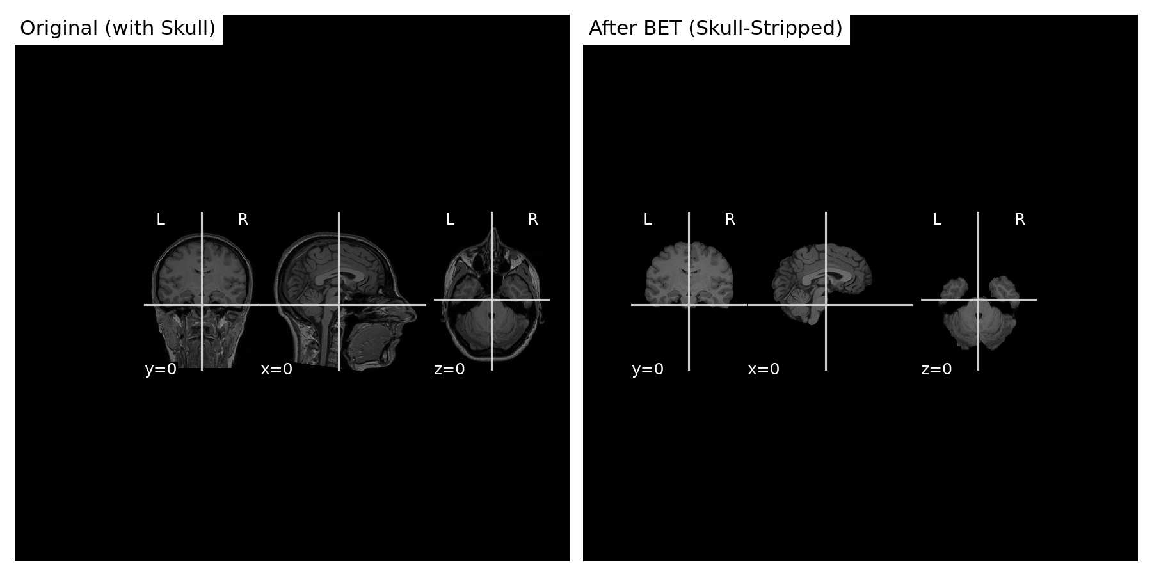}
\caption{An example of skull stripping using FSL BET.
The left-hand side image shows the original T1-weighted MRI with the skull and non-brain tissues. The right-hand side image shows the result after applying FSL BET. The brain region is cleanly extracted and the surrounding skull is removed.}
\label{fig:bet_comp}
\end{figure}

In Section \ref{sec:method}, we proposed the method of using FSL FAST and FSL BET to preprocess the images and create the dataset for training, validating and testing. Figure~\ref{fig:bet_comp} shows the result after applying FSL BET to the NIfTI format data and figure \ref{fig:fsl_result} below shows the result after applying FSL BET and FSL FAST.

\begin{figure}[H]
\centering
\includegraphics[width=0.5\textwidth]{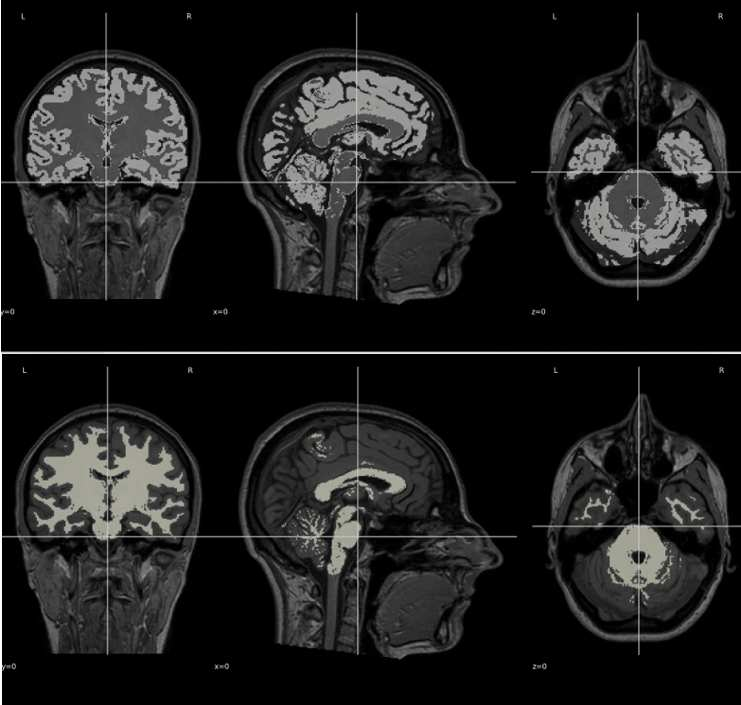}
\caption{An example of tissue segmentation using FSL FAST. The bottom set of images highlights the white matter, while the top set of images highlights the gray matter in the same subject. We have performed the segmentation on skull-stripped images. We have overlaid the tissue boundaries on coronal, sagittal, and axial slices. We have used these maps to construct pixelwise multi-class labels for model training.}
\label{fig:fsl_result}
\end{figure}

\begin{figure}[H]
\centering
\includegraphics[width=0.5\textwidth]{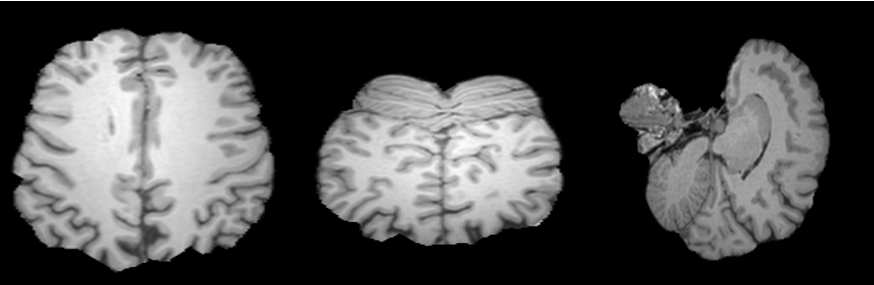}
\caption{Example slices from the axial plane, coronal plane, and sagittal plane (from left to right)}
\label{fig:slice_result}
\end{figure}

To evaluate segmentation performance, we tested each trained model on a randomly selected subset of 1,000 slices from the test dataset. This sampling strategy was employed to reduce evaluation time while still maintaining a fair comparison of performance across models.

\begin{table}[H]
\caption{Best overall Dice score for each model trained.}
\label{tab:dice_table}
\centering
\begin{tabular}{|c|c|}
\hline
Models & Best Overall Dice score \\
\hline
Axial model & 0.8717 \\
\hline
Sagittal model & 0.8604 \\
\hline
Coronal model & 0.8751 \\
\hline
Unified model (axial) & 0.8440 \\
\hline
Unified model (sagittal) & 0.8589 \\
\hline
Unified model (coronal) & 0.8742 \\
\hline
\end{tabular}
\end{table}

\begin{table}[H]
\caption{Best overall IoU score for each model trained.}
\label{tab:iou_table}
\centering
\begin{tabular}{|c|c|}
\hline
Models & Best Overall IoU score \\
\hline
Axial model & 0.7869 \\
\hline
Sagittal model & 0.7774 \\
\hline
Coronal model & 0.7935 \\
\hline
Unified model (axial) & 0.7601 \\
\hline
Unified model (sagittal) & 0.7750 \\
\hline
Unified model (coronal) & 0.7928 \\
\hline
\end{tabular}
\end{table}

Table \ref{tab:dice_table} and Table \ref{tab:iou_table} present the Dice coefficient and IoU scores for all trained models evaluated on a test set of 1000 randomly sampled slices. The coronal model achieved the highest performance with a Dice score of 0.8751 and IoU of 0.7935. The sagittal model ranked second, while the axial model showed the lowest performance among single-orientation models. For the unified model, it performed the best with the coronal dataset. Interestingly, the unified model trained on all three orientations did not perform better than the single-orientation models.

\begin{figure}[H]
\centering
\includegraphics[width=0.5\textwidth]{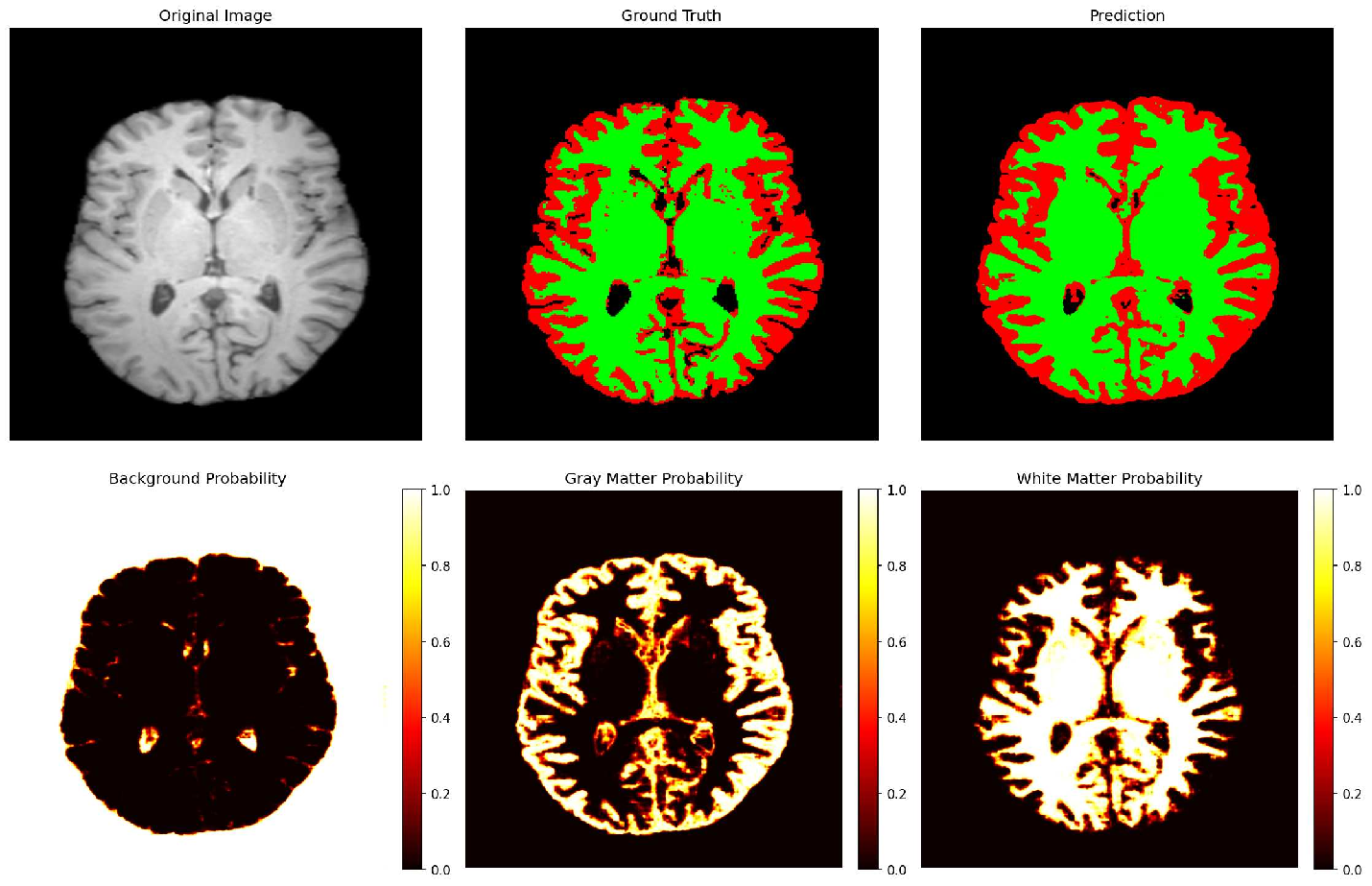}
\caption{Segmentation result from the axial model on an axial slice.}
\label{fig:ax_result_vis}
\end{figure}

\begin{figure}[H]
\centering
\includegraphics[width=0.5\textwidth]{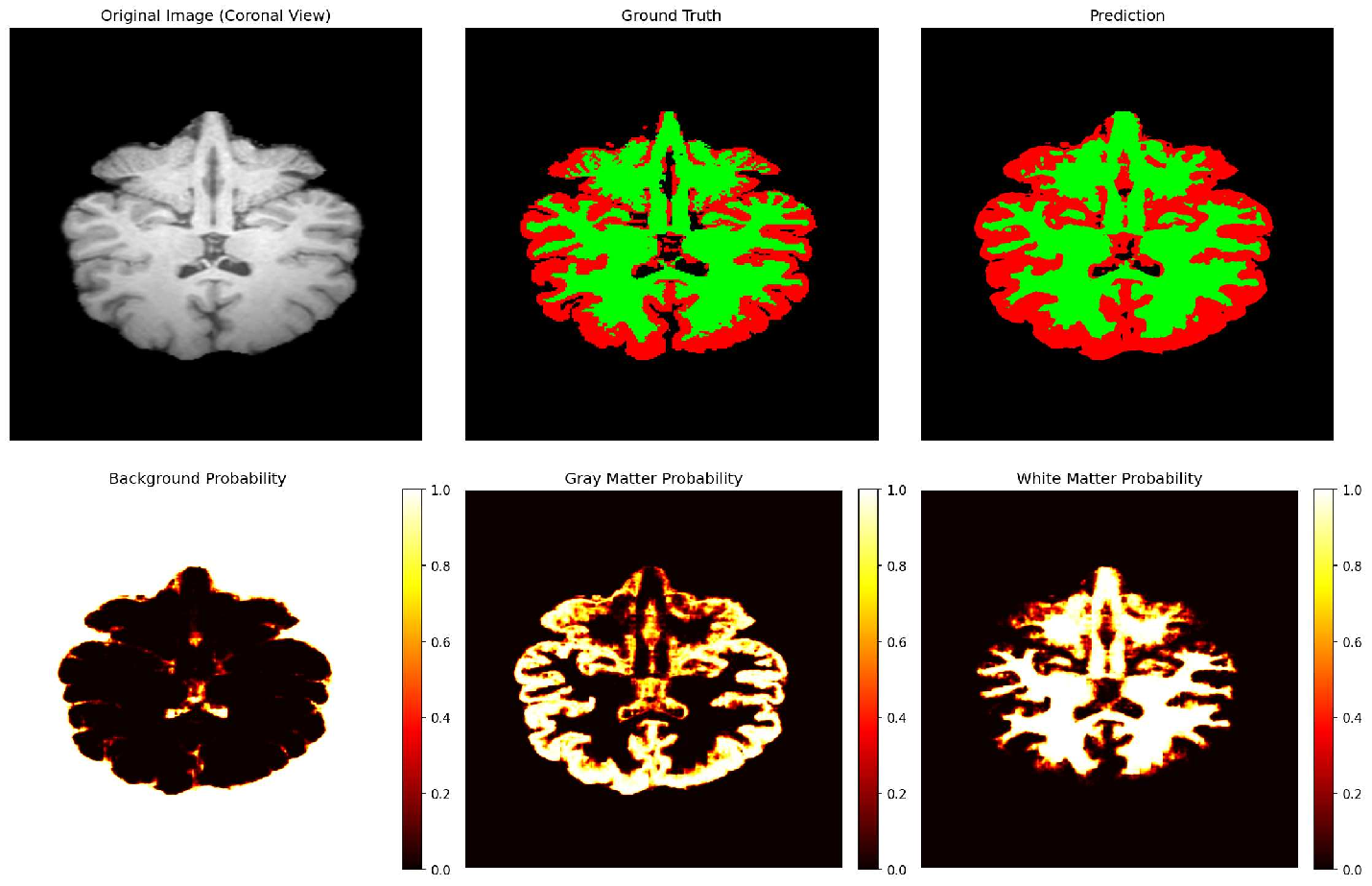}
\caption{Segmentation result from the coronal model on a coronal slice.}
\label{fig:cor_result_vis}
\end{figure}

\begin{figure}[H]
\centering
\includegraphics[width=0.5\textwidth]{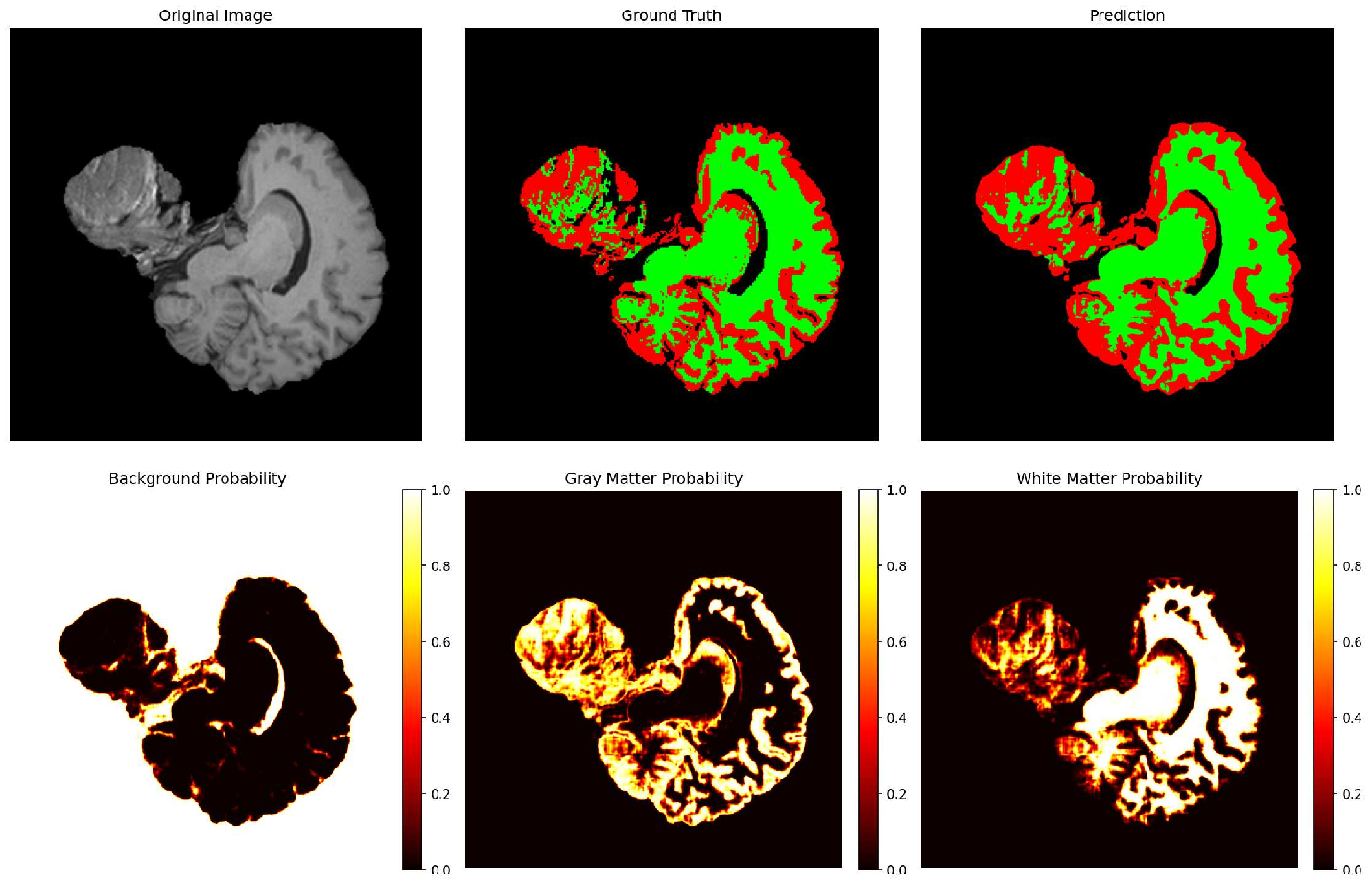}
\caption{Segmentation result from the sagittal model on a sagittal slice.}
\label{fig:sag_result_vis}
\end{figure}

\begin{figure}[H]
\centering
\includegraphics[width=0.5\textwidth]{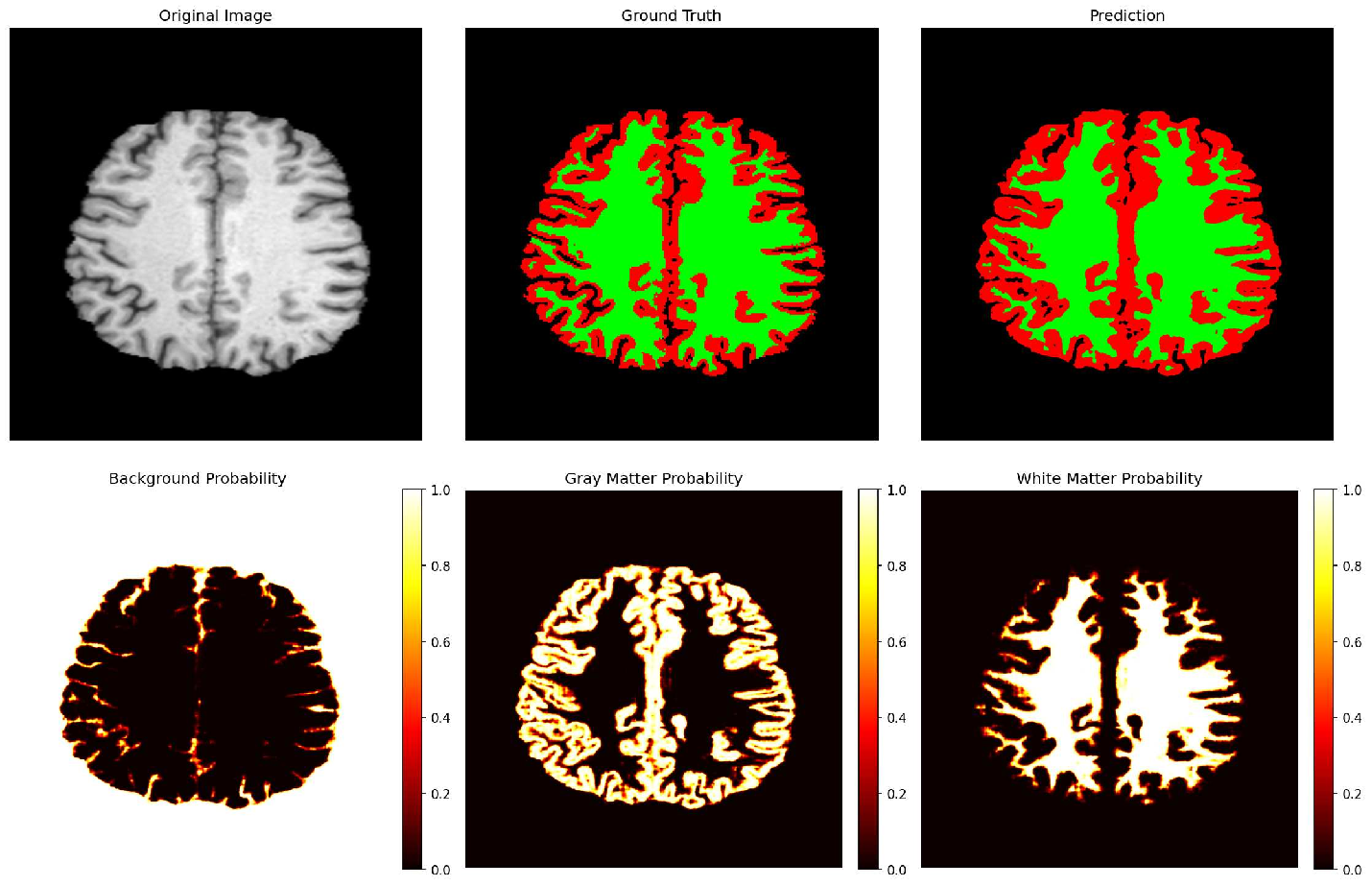}
\caption{Segmentation result from the unified model on an axial slice.}
\label{fig:uni_result_vis_ax}
\end{figure}

Figures \ref{fig:ax_result_vis}-\ref{fig:uni_result_vis_ax} present representative segmentation examples from each trained model. Each figure follows the following layout: the top row displays the original brain MRI slice (left), pseudo ground truth multi-class mask (middle), and model prediction (right). In all masks, red indicates gray matter, green indicates white matter, and black represents background. The bottom row shows predicted class probability maps from background, gray matter, and white matter as heatmaps scaled from 0.0 (darkest) to 1.0 (brightest), where brighter colors indicate higher model confidence.

\subsection{Discussions}

\subsubsection{Boundary Definition}
The performance of the coronal model suggests that anatomical orientation might impact segmentation quality. The coronal model demonstrates sharper boundaries between gray and white matter compared to other orientations, which may explain why the model performed better on coronal slices. The probability heatmaps show higher confidence values along tissue boundaries in the coronal model predictions.

\subsubsection{Model Consistency}
Comparing predictions across orientations, the coronal model appears to produce more consistent tissue classification, while the axial model occasionally misclassifies small gray matter regions as white matter (visible in the probability maps as ambiguous yellow regions).

\subsubsection{Confidence Patterns}
The probability maps reveal that all models exhibit high confidence in central brain regions but show uncertainty near the boundaries. This suggests that edge cases remain challenging even for fine-tuned foundation models.

\subsubsection{Unified Model Performance}
Contrary to our expectation, the unified model did not consistently outperform single-orientation models. This may be due to the increased variability in the training data, making it harder for the model to learn consistent features. The orientation-specific feature extraction has been diluted in multi-orientation training.

\subsubsection{Limitations}
We have derived the ground truth labels from FSL FAST probability maps rather than manual expert annotations, which is the key limitation. Since FSL FAST itself is a model-based segmentation tool rather than a gold standard, our reported Dice scores reflect similarity to FAST outputs rather than true anatomical accuracy. Therefore, the results should be interpreted as agreement with FAST-derived labels rather than absolute segmentation performance. The dataset that we used for training and evaluation was limited to the IXI dataset. We only used the T1-weighted MRI scans from healthy subjects. Although this dataset is well-structured and suitable for research purposes, it lacks diversity in terms of imaging protocols, scanner types, and subject demographics. In addition, it does not include cases with pathological conditions such as tumors, lesions, or neurodegenerative changes.

In this study, we excluded the slices containing minimal or background-only slices during preprocessing to reduce noise. This meant that the model did not learn how to confidently predict pure background cases. The model would become vulnerable when encountering slices with no brain tissue present.

\section{CONCLUSIONS}
\label{sec:conclusion}

This paper has presented a MedSAM-based approach for multi-class segmentation of brain MRI tissues, focusing on gray matter and white matter segmentation from T1-weighted scans. By modifying the MedSAM mask decoder and fine-tuning only the prompt encoder and decoder, we achieved competitive segmentation performance while preserving the pre-trained encoder's generalization capability.
Compared to traditional models like U-Net and nnU-Net, our approach highlights the potential of foundation models in medical imaging. MedSAM's prompt-based architecture requires fewer task-specific modifications. This makes our model highly adaptable across imaging modalities and anatomical structures.
However, our study is limited by the use of a single dataset of healthy subjects and 2D slice-based segmentation. Future work should explore 3D extensions of MedSAM, fine-tuning strategies for larger datasets, and performance evaluation on pathological cases.
In conclusion, this work demonstrates the viability of leveraging foundation models like Med-SAM for medical image segmentation, paving the way for future research on universal, adaptable segmentation systems in clinical applications.


\begin{thebibliography}{99}

\bibitem{katti2011magnetic}
Katti, G., Ara, S. A., and Shireen, A., ``Magnetic resonance imaging (MRI) -- A review,'' Int. J. Dental Clinics \textbf{3}(1), 65--70 (2011).

\bibitem{barha2016basics}
Barha, C. K., Nagamatsu, L. S., and Liu-Ambrose, T., ``Basics of neuroanatomy and neurophysiology,'' Handbook of Clinical Neurology \textbf{138}, 53--68 (2016).

\bibitem{mercadante2020neuroanatomy}
Mercadante, A. A. and Tadi, P., ``Neuroanatomy, gray matter,'' StatPearls Publishing (2020).

\bibitem{mao2018modeling}
Mao, D., Ding, Z., Jia, W., Liao, W., Li, G. J., Cao, H., He, Y., Ferdinando, H., Yeung, S., and Kwok, S., ``Modeling the differences in white and gray matter development,'' Med. Biol. Eng. Comput. \textbf{56}(9), 1579--1591 (2018).

\bibitem{woo2003regional}
Woo, M. A., Macey, P. M., Fonarow, G. C., Hamilton, M. A., and Harper, R. M., ``Regional brain gray matter loss in heart failure,'' J. Appl. Physiol. \textbf{95}(2), 677--684 (2003).

\bibitem{groh2025white}
Groh, A. M. R., Fournier, A. P., and Bhmann, A., ``White matter microstructure is associated with vascular and cardiac autonomic function,'' Cereb. Cortex \textbf{35}(2), bhae495 (2025).

\bibitem{ronneberger2015u}
Ronneberger, O., Fischer, P., and Brox, T., ``U-Net: Convolutional networks for biomedical image segmentation,'' Proc. MICCAI 2015, 234--241 (2015).

\bibitem{isensee2021nnu}
Isensee, F., Jaeger, P. F., Kohl, S. A., Petersen, J., and Maier-Hein, K. H., ``nnU-Net: A self-configuring method for deep learning-based biomedical image segmentation,'' Nat. Methods \textbf{18}(2), 203--211 (2021).

\bibitem{ma2024segment}
Ma, J., He, Y., Li, F., Han, L., You, C., and Wang, B., ``Segment anything in medical images,'' Nat. Commun. \textbf{15}, 654 (2024).

\bibitem{smith2002fast}
Smith, S. M., ``Fast robust automated brain extraction,'' Hum. Brain Mapp. \textbf{17}(3), 143--155 (2002).

\bibitem{zhang2002segmentation}
Zhang, Y., Brady, M., and Smith, S., ``Segmentation of brain MR images through a hidden Markov random field model and the expectation-maximization algorithm,'' IEEE Trans. Med. Imaging \textbf{20}(1), 45--57 (2001).

\bibitem{tajbakhsh2016convolutional}
Tajbakhsh, N., Shin, J. Y., Gurudu, S. R., Hurst, R. T., Kendall, C. B., Gotway, M. B., and Liang, J., ``Convolutional neural networks for medical image analysis: Full training or fine tuning?,'' IEEE Trans. Med. Imaging \textbf{35}(5), 1299--1312 (2016).

\bibitem{ahmed2008}
Ahmed, M. N. and Yamany, S. M., ``A modified fuzzy C-means algorithm for bias field estimation and segmentation of MRI data,'' IEEE Trans. Med. Imaging \textbf{21}(3), 193--199 (2002).

\bibitem{pan2007regiongrowing}
Pan, Z. and Lu, J., ``A Bayes-based region-growing algorithm for medical image segmentation,'' Comput. Sci. Eng. \textbf{9}(4), 32--38 (2007).

\bibitem{abras2005kmeans}
Abras, C. N., ``Brain MRI segmentation using K-means clustering,'' J. Digit. Imaging \textbf{18}(4), 339--345 (2005).

\bibitem{ng2006clustering}
Ng, H. P., Ong, S. H., Foong, K. W. C., Goh, P. S., and Nowinski, W. L., ``Medical image segmentation using K-means clustering and improved watershed algorithm,'' Proc. IEEE Southwest Symp. Image Anal. Interpretation, 61--65 (2006).

\bibitem{salem2013review}
Salem, S. A. and Salem, N. M., ``A review on brain MRI image segmentation techniques,'' Int. J. Comput. Appl. \textbf{84}(9), 1--10 (2013).

\bibitem{wu2025survey}
Wu, J., et al., ``A survey on deep learning for medical image segmentation,'' Neurocomputing (2025).

\bibitem{kamnitsas2017efficient}
Kamnitsas, K., Ledig, C., Newcombe, V. F., Simpson, J. P., Kane, A. D., Menon, D. K., Rueckert, D., and Glocker, B., ``Efficient multi-scale 3D CNN with fully connected CRF for accurate brain lesion segmentation,'' Med. Image Anal. \textbf{36}, 61--78 (2017).

\bibitem{kirillov2023segment}
Kirillov, A., Mintun, E., Ravi, N., Mao, H., Rolland, C., Gustafson, L., Xiao, T., Whitehead, S., Berg, A. C., Lo, W.-Y., Dollar, P., and Girshick, R., ``Segment anything,'' Proc. IEEE/CVF ICCV, 4015--4026 (2023).

\bibitem{raghu2019transfusion}
Raghu, M., Zhang, C., Kleinberg, J., and Bengio, S., ``Transfusion: Understanding transfer learning for medical imaging,'' Proc. NeurIPS, 3347--3357 (2019).

\bibitem{dosovitskiy2020image}
Dosovitskiy, A., Beyer, L., Kolesnikov, A., Weissenborn, D., Zhai, X., Unterthiner, T., Dehghani, M., Minderer, M., Heigold, G., Gelly, S., Uszkoreit, J., and Houlsby, N., ``An image is worth 16x16 words: Transformers for image recognition at scale,'' Proc. ICLR (2021).

\bibitem{loshchilov2019decoupledweightdecayregularization}
Loshchilov, I. and Hutter, F., ``Decoupled weight decay regularization,'' Proc. ICLR (2019).

\end{thebibliography}
\end{document}